\documentclass[conference]{IEEEtran}
\usepackage{graphicx} 
\usepackage{amsmath} 
\usepackage{hyperref} 

\begin{document}

\title{Advancing Autonomous Driving Perception:\\Analysis of Sensor Fusion and Computer Vision Techniques}

\author{
\IEEEauthorblockN{Urvishkumar Bharti}
\IEEEauthorblockA{MSc Robotics, ECE Department\\
Drexel University\\
Philadelphia, PA, USA\\
Email: ub48@drexel.edu}
\and
\IEEEauthorblockN{Vikram Shahapur}
\IEEEauthorblockA{MSc Robotics, ECE Department\\
Drexel University\\
Philadelphia, PA, USA\\
Email: vs642@drexel.edu}

}

\maketitle

\begin{abstract}
In autonomous driving, perception systems are pivotal as they interpret sensory data to understand the environment, which is essential for decision-making and planning. Ensuring the safety of these perception systems is fundamental for achieving high-level autonomy, allowing us to confidently delegate driving and monitoring tasks to machines. This report aims to enhance the safety of perception systems by examining and summarizing the latest advancements in vision based systems, and metrics for perception tasks in autonomous driving. The report also underscores significant achievements and recognized challenges faced by current research in this field. This project focuses on enhancing the understanding and navigation capabilities of self-driving robots through depth based perception and computer vision techniques. Specifically, it explores how we can perform better navigation into unknown map 2D map with existing detection and tracking algorithms and on top of that how depth based perception can enhance the navigation capabilities of the wheel based bots to improve autonomous driving perception.
\end{abstract}

\section{Introduction}
Autonomous driving systems rely heavily on accurate and robust perception of the environment. This project investigates two key approaches: challenges into sensor fusion techniques, and aiming to enhance vision based navigation and decision-making capabilities of self-driving vehicles.
\subsection{Literature Review}
Michel Devy, outlines a task-focused strategy devised for the perception system of an autonomous robot designed for cross-country navigation. They propose an adaptive navigation technique, which is particularly effective in dealing with the complexities of natural environments. They have introduced a comprehensive, multi-tiered perception system for an autonomous robot designed for cross-country navigation. This system highlights a variety of modeling services, significantly boosting the robot's autonomy and efficiency [5].  Benjamin Ranft1 introduces a method for autonomous navigation tailored for Micro Air Vehicles, particularly cost-effective models. This method depends solely on a single camera and a handful of additional onboard sensors to address the issues of flight planning and collision avoidance. Despite the absence of a direct 3D sensor, their system can deduce metric distances from its monocular camera using two complementary techniques. One of these involves overlaying an oscillating motion pattern onto the regular flight path to accurately determine the current 3D positions of sparse image features. They have detailed every component of a comprehensive system that has demonstrated its ability for autonomous indoor navigation. Even without a built-in 3D sensor on their quad-copter, it can carry out metric 3D reconstructions primarily using a single-lens camera [6]. Kunyan Zhu, introduces an autonomous navigation technique for robots that utilizes a multi-camera configuration to benefit from a broad field of view. They have developed a novel multi-task network that processes visual data from the left, center, and right cameras. This network is capable of identifying navigable areas, detecting intersections, and deducing steering directions which enhances the robot’s ability
 of path planning and obstacle avoidance [7].
 
 \begin{figure}[h]
    \centering
    \includegraphics[width=1\columnwidth]{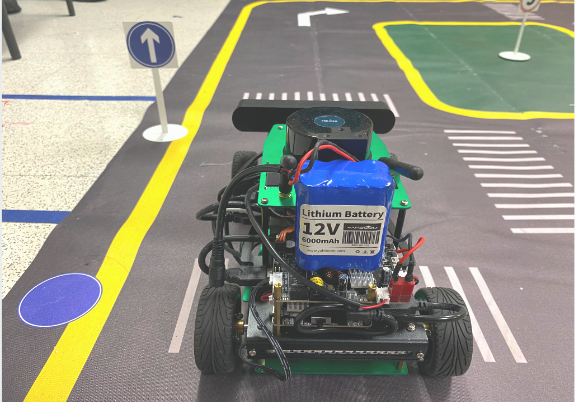}
    \caption{Rosmaster R2 performing Autonomous Navigation} 
\end{figure}

Suraj Bijjahalli's study offers an exhaustive evaluation of traditional UAS navigation systems, encompassing elements like system structure, sensor types, and data integration algorithms. It also critically examines and measures performance monitoring and enhancement strategies against existing and prospective UAS Traffic Management (UTM) standards. The main emphasis is on pinpointing significant gaps in existing literature where the application of AI-based techniques could potentially improve navigation performance [8]. Yusras and his team conducted a comprehensive review of the latest techniques in Visual Odometry (VO) and Visual-Inertial Odometry (VIO). They also examined research related to localization in environments with visual impairments. The VO techniques and associated studies were scrutinized based on critical design elements such as appearance, feature, and learning-based methodologies. Conversely, VIO-related research was classified according to the extent and nature of the fusion process into loosely coupled, semi-tightly coupled, or tightly-coupled approaches, and filtering or optimization-based paradigms [9].In an effort to facilitate reproducible assessments of social navigation algorithms, Nathan Tsoi introduced the Social Environment for Autonomous Navigation (SEAN). SEAN is an open-source, expandable social navigation simulation platform with high visual accuracy. It comes with a toolkit specifically designed for evaluating navigation algorithms. We showcase the capabilities of SEAN and its evaluation toolkit in two distinct environments featuring dynamic pedestrians and utilizing two different robots [11].

\section{Background}
In this project, we utilized depth-based perception to enable autonomous navigation of the robot in an unfamiliar environment. The fusion of 2D LiDAR and depth camera sensors demanded substantial computational resources, leading to system throttle errors during the object detection task alone. In addition to object detection, we also maneuvered the Rosmaster R2 bot autonomously, detecting traffic signs such as 'Move', 'Turn', and 'Stop'.
Depth cameras and traditional cameras play critical roles in mobile robot perception, providing 3D environmental information and facilitating vision-guided navigation, respectively. Fig1  shows such example of the camera that we have used in this project.

 \begin{figure}[h]
    \centering
    \includegraphics[width=1\columnwidth]{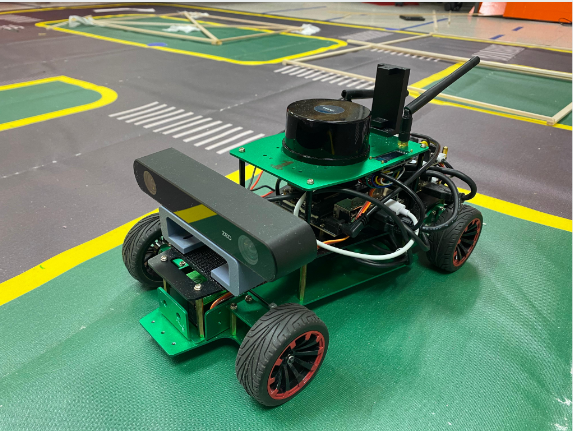}
    \caption{ROSMASTER R2} 
\end{figure}

\section{Methodology}
This section elaborates on the comprehensive methodology adopted for enhancing autonomous driving perception through depth-based perception.

\subsection{Hardware and Software Setup}
The project utilized a combination of advanced hardware and software to process and analyze sensor data:

\begin{itemize}
    \item \textbf{Jetson Xavier Processor:} Served as the computational backbone, handling data processing and model execution.
    \item \textbf{ZED 2 RGBD Camera:} Provided high-resolution images and depth data, crucial for object detection and distance estimation.  Fig 2  shows such example of the camera that we have used in this project.
    \item \textbf{ROS (Robot Operating System):} Enabled efficient system integration, data handling, and algorithm implementation.
    \item \textbf{ZED SDK:} Offered tools and APIs for extracting and processing data from the ZED 2 camera.
\end{itemize}

The integration of these hardware components through ROS facilitated a modular approach, allowing for the independent development and testing of subsystems.

\begin{figure}[h]
    \centering
    \includegraphics[width=1\columnwidth]{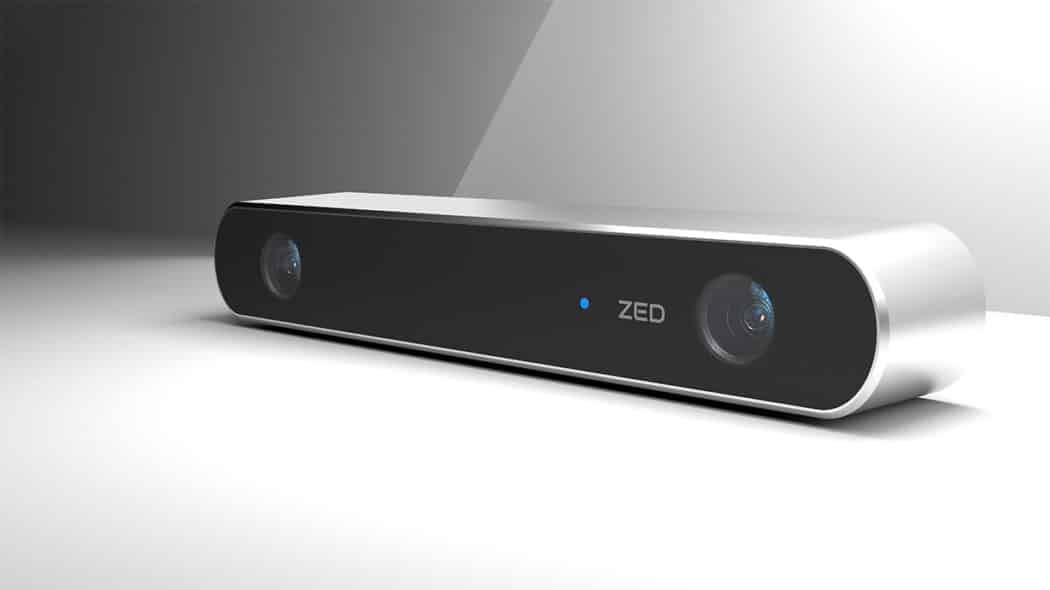}
    \caption{ZED2 RGBD Camera} 
\end{figure}

\subsection{Sensor Fusion Technique}
 Performing sensor fusion is a computationally intensive task, particularly in the context of autonomous navigation systems. Sensor fusion is an essential component of many perception systems, such as autonomous driving and robotics. It involves the integration of data from multiple sensors to provide a more accurate understanding of the environment such as LiDAR and RGB Cameras. This process requires significant computational resources due to the complexity of the algorithms used for data integration and the large volume of data generated by the sensors.

 In this section, we outline how the sensor fusion algorithm works in estimating the distance to a traffic sign detected by the YOLOv5 model using data from a 2D lidar and camera. The algorithm leverages either the Extended Kalman Filter (EKF) or the Unscented Kalman Filter (UKF) to integrate sensor measurements and predict the state of the traffic sign.

\subsubsection{State Vector}
The state vector representing the position and velocity of the traffic sign is defined as:
\[ \mathbf{x} = \begin{bmatrix} x \\ y \\ v_x \\ v_y \end{bmatrix}, \]
where \( (x,y) \) represents the position and \( (v_x, v_y) \) represents the velocity.

\subsubsection{Measurement Model}
Lidar data is used to measure the position directly within the lidar's coordinate frame. Camera data, obtained through YOLOv5 detections, provide bounding box coordinates, which are transformed to align with lidar measurements.

\subsubsection{Prediction Step}
Applicable to both Extended Kalman Filter (EKF) and Unscented Kalman Filter (UKF), the constant velocity model is employed:
\[ \mathbf{x}_{k+1} = \mathbf{F} \cdot \mathbf{x}_k, \]
where \( \mathbf{F} \) is the state transition matrix.

\subsubsection{Update Step}

{Extended Kalman Filter:}
\begin{enumerate}
    \item Calculate the Jacobian matrix \( \mathbf{H} \) of the measurement model.
    \item Compute the Kalman Gain \( \mathbf{K} \) using predicted state covariance \( \mathbf{P} \), measurement noise covariance \( \mathbf{R} \), and Jacobian matrix \( \mathbf{H} \).
    \item Update the state estimate and covariance using the Kalman Gain and the residual.
\end{enumerate}

{Unscented Kalman Filter:}
\begin{enumerate}
    \item Generate sigma points from the predicted state and covariance.
    \item Propagate sigma points through the nonlinear measurement model to obtain predicted measurements.
    \item Compute the mean and covariance of the predicted measurements.
    \item Calculate the Kalman Gain using the predicted and measured covariance matrices.
    \item Update the state estimate using the Kalman Gain and the difference between predicted and measured values.
\end{enumerate}

\subsubsection{Initialization}
Initialize the state vector, state covariance matrix, process noise covariance matrix, and measurement noise covariance matrix as required.

\subsubsection{Integration and Optimization}
Integrate the sensor fusion algorithm within the ROS environment for compatibility and efficiency, ensuring real-time processing capabilities on the Jetson Xavier NX processor.

For instance, the FUTR3D framework, a unified sensor fusion framework for 3D detection, can be used in almost any sensor configuration. It employs a query-based Modality-Agnostic Feature Sampler (MAFS), together with a transformer decoder with a set-to-set loss for 3D detection. This avoids using late fusion heuristics and post-processing tricks, but it requires substantial computational power.

Moreover, the performance of multiple integrated sensors can directly determine the safety and feasibility of automated driving vehicles. Therefore, the computational power needed for sensor fusion is not only a requirement but also a critical factor in the successful implementation of autonomous navigation systems.

\subsection{Challenges and Solution}

The project faced significant challenges in terms of resource and time constraints. The computational capacity of the Jetson Xavier was limited, which posed a significant hurdle. The team was unable to implement 2D LiDAR fusion and the ZED 2 camera as initially planned due to these constraints. Additionally, the project demanded significant computational power for processing depth information from the ZED2 RGBD camera and running the YOLOv5 object detection model simultaneously to perform Autonomous Navigation. These computation demands added to the complexity and challenges of the project.

Despite the challenges, the team managed to devise effective solutions and achieve significant milestones. For depth-based perception, the ZED 2 camera was utilized for depth and RGB data. The team estimated distances to objects using the depth map, which proved to be a valuable asset for the project. For autonomous navigation, the team incorporated computer vision techniques to detect lanes and control the robot accordingly. We relied on depth data to get the distance of the object (traffic sign), which was crucial for the navigation system. To address the resource constraints, the team optimized algorithms for computational efficiency. We achieved real-time performance on Jetson Xavier by prioritizing system efficiency, which significantly improved the overall performance of the project.

\subsection{Working}
\subsubsection{Lane Detection}
In this project, we embarked on a journey to develop an autonomous navigation system, starting with the fundamental task of detecting lanes. The lane-following functionality of the robot was achieved through a sophisticated combination of computer vision techniques and control algorithms. This ensured precise navigation along road lanes. Through the integration of color filtering, the robot adeptly identified and tracked lane markings, enabling it to maintain a stable and centered position within the lanes. The image processing pipeline was designed to process and analyze the visual data captured by the robot's sensors. The first step was the application of a Gaussian Blur, followed by a Yellow Color Filtering and Yellow Color Mask to highlight the lane markings in the image. Finally, Region of Interest Masking was used to focus the robot's attention on the relevant areas of the image where lane markings are likely to be found.

\begin{figure}[h]
    \centering
    \includegraphics[width=1\columnwidth]{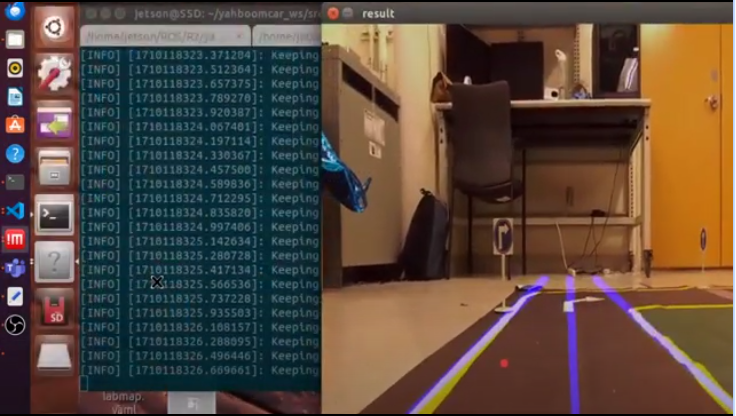}
    \caption{Lane Detection} 
\end{figure}

\subsubsection{Object Detection} We utilized the YOLOv5 pre-trained model for object detection. This model has been widely used in various applications, including lane detection, missing road lane markings detection, and pedestrian detection. The use of YOLOv5 allowed us to effectively detect objects in real-time, contributing significantly to the success of the project. Here we have detected successfully the 'Stop', 'Move', and 'Turn' signs using which robot will perform the task according to the signs. One such example of a lane detection algorithm is shown in Fig 3. The system was designed to detect lane markings and employed robust algorithms for tracking and control. This ensured smooth, precise navigation along the road. The control algorithms were designed to respond dynamically to changes in the environment, allowing the robot to adapt to varying road conditions and maintain a steady course. The Hough Line Transform was used to detect straight lines in the image. It processed the detected edges in the image and identified line segments as lane markings. This information was then used to guide the robot's navigation and ensure it stayed within the lanes. The system calculated the midpoint between the detected lane boundaries to determine the robot's position within the lane. This information was crucial for the lane-centering functionality, which aimed to keep the robot centered within the lane to ensure safe and efficient navigation. During ROS Integration, the module publishes two types of information to ROS topics, "detected-class": This topic publishes the names of the detected objects. And "detected-class-distance": This topic publishes the centroid coordinates of the detected objects. The module processes images to extract bounding boxes, confidence scores, and class labels. If the confidence score of detection is greater than 0.8, the module publishes the class name and centroid of the detected object. The module visualizes the detection results by displaying bounding boxes, class names, confidence scores, and centroids on the images. This visualization aids in understanding the performance and accuracy of the object detection module. 

\begin{figure}[h]
    \centering
    \includegraphics[width=0.9\columnwidth]{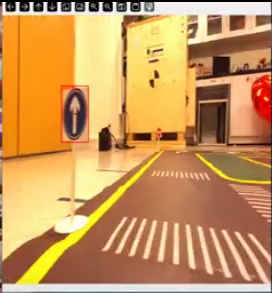}
    \caption{YOLO Model Detecting the Signs} 
\end{figure}

\subsubsection{Depth Estimation}The ZED 2 camera is a powerful stereo camera that plays a crucial role in depth estimation for autonomous navigation. It combines advances in AI, sensor hardware, and stereo vision to build an unmatched solution in spatial perception and understanding. The camera features ultra-wide depth perception with a 110-degree horizontal and 70-degree vertical field of view, including optical distortion compensation. It also has enhanced low-light vision with an f/1.8 aperture and improved ISP, capturing 40 percent more light in dark environments. The ZED 2 camera uses stereo vision and neural networks to replicate human-like vision, enabling depth perception from 0.2 to 20m². This depth perception capability is essential for autonomous navigation as it allows the system to understand the 3D structure of the environment. For depth estimation, the ZED 2 camera uses a depth map, which is a 2D representation where each pixel corresponds to the distance from the camera to the corresponding point in the environment. This depth map is generated by comparing the images from the two lenses of the ZED 2 camera and estimating the distance to each pixel based on the difference in position of that pixel in the two images.

\begin{figure}[h]
    \centering
    \includegraphics[width=1\columnwidth]{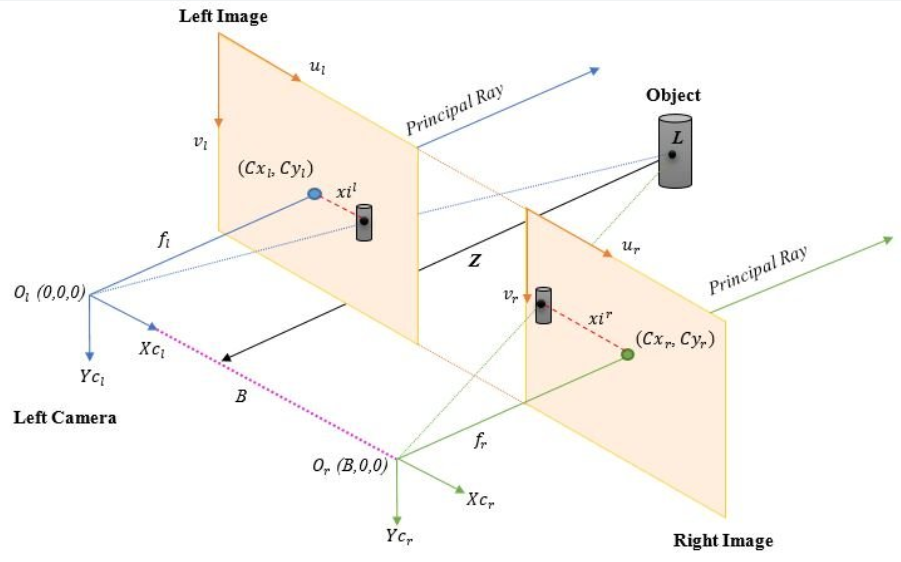}
    \caption{Stereo Vision} 
\end{figure}

In autonomous navigation, this depth information is used to identify obstacles, plan paths, and make decisions about the robot's movements. For instance, the depth map can be used to identify areas that are too steep or too rough for the robot to traverse. It can also be used to estimate the distance to a target or to other vehicles or pedestrians, which is crucial for collision avoidance.

\subsubsection{PID Controller}
The motion of a differential drive robot can be described by its linear velocity ($v$) and angular velocity ($\omega$), which are related to the velocities of the right ($v_r$) and left wheels ($v_l$) as follows:

\[
v = \frac{v_r + v_l}{2}
\]
\[
\omega = \frac{v_r - v_l}{L}
\]

where $L$ is the distance between the two wheels.

\subsubsection*{Application of PID Control}

For lane following, a PID controller adjusts $v$ and $\omega$ to minimize the error between the robot's current position and the desired path. The control inputs for linear and angular velocities are given by:

\[
u_v = K_{p_v} e_v + K_{i_v} \int e_v \, dt + K_{d_v} \frac{de_v}{dt}
\]
\[
u_\omega = K_{p_\omega} e_\omega + K_{i_\omega} \int e_\omega \, dt + K_{d_\omega} \frac{de_\omega}{dt}
\]

where $u_v$ and $u_\omega$ are the control inputs for linear and angular velocities, respectively; $e_v$ and $e_\omega$ are the errors in linear and angular velocities; and $K_{p_v}$, $K_{i_v}$, $K_{d_v}$, $K_{p_\omega}$, $K_{i_\omega}$, and $K_{d_\omega}$ are the PID coefficients for linear and angular control, respectively.
\subsubsection{Linearized Model}

Let's consider the linearized model of a robot's kinematics as follows:

\[
A_{t-1} = 
\begin{bmatrix}
\frac{\partial f_1}{\partial x_{t-1}} & \frac{\partial f_1}{\partial y_{t-1}} & \frac{\partial f_1}{\partial \psi_{t-1}} \\
\frac{\partial f_2}{\partial x_{t-1}} & \frac{\partial f_2}{\partial y_{t-1}} & \frac{\partial f_2}{\partial \psi_{t-1}} \\
\frac{\partial f_3}{\partial x_{t-1}} & \frac{\partial f_3}{\partial y_{t-1}} & \frac{\partial f_3}{\partial \psi_{t-1}} \\
\end{bmatrix}
\]

Where \( x_{t-1} \), \( y_{t-1} \), and \( \psi_{t-1} \) denote the robot's position on the \(x\) and \(y\) axes, and its rotation in radians at the previous time step, respectively.

\subsubsection{Observation Model}

The observation model can be expressed as:

\[
O_t = H_t x_t + w_t
\]

Where:
\begin{itemize}
    \item \(O_t\) is the observed state.
    \item \(H_t\) is the observation matrix.
    \item \(x_t\) is the state vector at time \(t\).
    \item \(w_t\) represents the added sensor noise.
\end{itemize}

The observation vector is given by:

\[
\begin{bmatrix}
O_1 \\
O_2 \\
O_3 \\
\end{bmatrix}
=
H_t
\begin{bmatrix}
x_t \\
y_t \\
\psi_t \\
\end{bmatrix}
+
\begin{bmatrix}
w_1 \\
w_2 \\
w_3 \\
\end{bmatrix}
\]

\subsubsection{Range and Bearing Equations}

Using trigonometry, we can derive the range (\(r\)) and bearing (\(b\)) to a landmark as:

\[
r = \sqrt{(x_t - x_{landmark})^2 + (y_t - y_{landmark})^2}
\]
\[
b = \text{atan2}(y_{landmark} - y_t, x_{landmark} - x_t)
\]

The range and bearing in matrix form:

\[
\begin{bmatrix}
r \\
b \\
\end{bmatrix}
=
\begin{bmatrix}
h_1 \\
h_2 \\
\end{bmatrix}
\]

Linearization of the sensing model involves computing the Jacobian matrix \(H_t\) of the measurement function with respect to the state vector:

\[
H_t =
\begin{bmatrix}
\frac{\partial r}{\partial x_{t-1}} & \frac{\partial r}{\partial y_{t-1}} & \frac{\partial r}{\partial \psi_{t-1}} \\
\frac{\partial b}{\partial x_{t-1}} & \frac{\partial b}{\partial y_{t-1}} & \frac{\partial b}{\partial \psi_{t-1}} \\
\end{bmatrix}
\]
The system utilized a PID (Proportional-Integral-Derivative) controller for dynamic steering adjustment. The PID controller adjusted the robot's steering based on the deviation from the desired lane position. This allowed the robot to make smooth and precise adjustments to its course, ensuring it stayed on track and navigated the lanes effectively. Throughout this project, we tried to improve our control as much as possible resulting the smooth navigation within the environment.

Update Function:
    The update() method of the PID controller takes the measured value as input and returns the control output. Here's how it works:
        Proportional Term: This term is calculated by multiplying the proportional gain (Kp) with the current error, which is the difference between the desired setpoint and the measured value.
        Integral Term: The integral of the error is accumulated over time and multiplied by the integral gain (Ki). This helps in eliminating steady-state error and responding to sustained deviations from the setpoint.
        Derivative Term: This term is proportional to the rate of change of the error and is calculated as the product of derivative gain (Kd) and the difference between the current error and the previous error. It helps in reducing overshoot and dampening oscillations.
        Control Output: The control output is the sum of the proportional, integral, and derivative terms.

    Usage in the Code:
        The PID controller is instantiated with specific values for Kp, Ki, Kd, and setpoint.
        In the process-image() function, the output of the PID controller is used to adjust the angular velocity of the robot based on the deviation of the detected lane from the center.
        The deviation (center offset) is calculated based on the position of the detected lane relative to the center of the image.
        The PID controller's update() method is called with center-offset as the measured value to compute the control output.
        The control output is then used to adjust the robot's angular velocity (twist.angular.z) to keep the robot aligned with the center of the lane.

    Parameters:
        Kp, Ki, and Kd: These parameters are crucial for tuning the PID controller's response. They determine the balance between the proportional, integral, and derivative control actions. Adjusting these values affects the controller's performance in terms of stability, responsiveness, and steady-state error.
        setpoint: This is the desired value that the controller aims to achieve. In this case, it represents the center of the lane.
        center-offset: This is the error term calculated based on the deviation of the detected lane from the center. It serves as the input to the PID controller.

In summary, the PID controller in this code adjusts the robot's angular velocity based on the deviation of the detected lane from the center, helping the robot maintain its position within the lane.

The process for navigation involved the following steps:

\begin{enumerate}
    \item \textbf{Data Synchronization:} Time-stamping data streams from sensor RGBD to ensure alignment. Key computer vision techniques were employed for lane detection, object recognition, and distance estimation.
    \item \textbf{Calibration:} Spatially aligning the RGBD camera data using calibration matrices.
    \item \textbf{Data Integration:} Merging RGBD images with Control commands to create a unified environmental model.
\end{enumerate}

This approach allowed the robot to leverage the strengths of each sensor type, improving navigation and obstacle detection capabilities.

\subsection{ROS Integration}
The Robot Operating System (ROS) played a pivotal role in the implementation of this project. ROS is a flexible framework for writing robot software and provides services designed for a heterogeneous computer cluster such as hardware abstraction, low-level device control, implementation of commonly-used functionality, message-passing between processes, and package management.

In the context of lane following, ROS provided the necessary infrastructure for interfacing with the camera and processing the images. The images captured by the camera were published as ROS messages, which were then subscribed to by the lane detection node. The lane detection node applied color filtering and line fitting techniques to identify the lanes and published the detected lanes as ROS messages. The control node subscribed to these messages and adjusted the robot's steering to stay within the lanes.

For object detection, the YOLOv5 model was integrated into the ROS framework. The images from the camera were passed to the YOLOv5 node, which detected objects of interest and published the detections as ROS messages. These messages were used to trigger specific actions by the robot.

The ZED 2 camera was used to obtain depth information, which was crucial for object distance estimation. The depth information was published as ROS messages, which were then used to estimate the distance to detected objects. This information was crucial for making decisions about when to execute certain actions, such as stopping or turning.

The control of the robot was implemented using a PID controller, which was integrated into the ROS framework. The PID controller adjusted the robot's steering based on the deviation from the desired lane position, ensuring that the robot remained centered within the lane.

\begin{figure}[h]
    \centering
    \includegraphics[width=1\columnwidth]{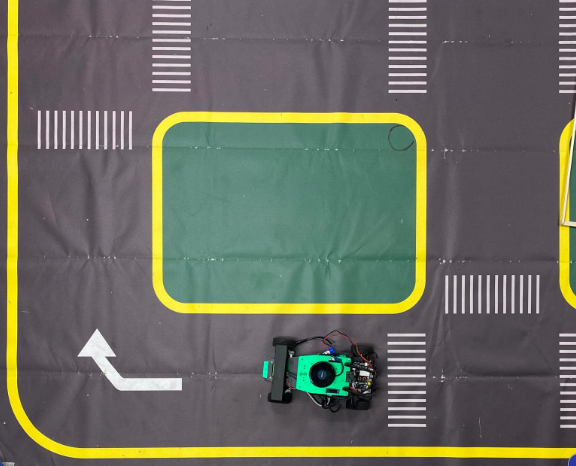}
    \caption{Implementation Environment} 
\end{figure}
\subsection{Results and Analysis}

The robot demonstrated a high degree of proficiency in following lane markings detected in the camera images. It employed techniques such as color filtering and line fitting to identify lanes. The robot was able to adjust its steering to stay within the lanes, demonstrating the effectiveness of the control algorithms implemented. The success in lane following is a testament to the robustness of the computer vision techniques and control algorithms used in this project.

The robot was capable of detecting objects of interest within its environment, such as "Turn", "Go", or "Stop" signs. This was achieved using advanced computer vision techniques. The detection was used to trigger specific actions by the robot, adding a layer of interactivity and responsiveness to the system. The successful detection of objects of interest and the subsequent triggering of actions demonstrate the effectiveness of the object detection algorithms implemented.

The robot was able to estimate the distance to detected objects using depth information obtained from the camera. This information was crucial for making decisions about when to execute certain actions, such as stopping or turning. The ability to accurately estimate distances to objects is a critical component of autonomous navigation systems, as it allows the system to interact safely and effectively with its environment. The success in object distance estimation indicates the effectiveness of the depth perception capabilities of the system.

\section{Conclusion}

In conclusion, the results obtained from this project demonstrate the effectiveness of the techniques and algorithms implemented. The robot was able to follow lanes, detect objects, and estimate distances to objects accurately, demonstrating its potential for real-world applications in autonomous navigation.

This project demonstrated the effective use of computer vision techniques, control algorithms, and the YOLOv5 model in developing an autonomous navigation system. Despite the challenges faced, the project was successful in achieving its objectives and contributed to the field of autonomous navigation.

\section*{Acknowledgment}
The authors would like to thank Dr. Lifeng Zhou and Zhou Labs for their guidance and support throughout this project.

\section{Contribution}
Vikram Shahpur: Lane Tracking and Control, Object Distance Estimation, Depth Data Processing

Urvishkumar Bharti: Lane Following, Object Distance Estimation/Image Processing, Lane Tracking

\end{document}